%% file: main.tex
\documentclass{llncs}
\usepackage{enumerate}
\usepackage{hyperref}
\usepackage{amsmath}
\usepackage{amssymb}     
\usepackage{latexsym} 
\usepackage[tableposition=top,justification=justified,font={scriptsize}]{caption}
\usepackage{graphics}    
\usepackage{graphicx}
\usepackage{pgfplots}
\usepackage{tikz}
\DeclareMathAlphabet{\mathcalligra}{T1}{calligra}{m}{n}
\usetikzlibrary{arrows,petri,topaths,backgrounds,snakes,patterns,positioning}
\usepackage{tkz-berge}
\usetikzlibrary{pgfplots.groupplots}
\usepackage{subfig}
\usepackage{xcolor}
\usepackage{float}
\usepackage{algorithmic}
\usepackage[linesnumbered,lined,boxed]{algorithm2e}
\usepackage{ragged2e}
\usepackage{tabularx} 
\usepackage{booktabs}
\usepackage[flushleft]{threeparttable}
\usepackage[labelfont=bf]{caption}
\usepackage{rotating}
\usepackage{multirow}
\SetAlCapSkip{1em}

\newcommand{\myparagraph}[1]{\vspace*{1mm}\noindent {\bf #1}}

\newcommand{\reals}{\mathbb{R}}

\newcommand{\abs}[1]{\left|#1\right|}

\usepackage{wrapfig}
\title{Fast Sequence Based Embedding with Diffusion Graphs}

\author{Benedek Rozemberczki \and Rik Sarkar}

\institute{School of Informatics, University of Edinburgh, U.K.
\email{benedek.rozemberczki@ed.ac.uk\\ rsarkar@inf.ed.ac.uk}}
\begin{document}
\maketitle
\begin{abstract}
A graph embedding is a representation of graph vertices in a low dimensional space, which approximately preserves properties such as distances between nodes. Vertex sequence based embedding procedures use features extracted from linear sequences of nodes to create embeddings using a neural network. In this paper, we propose diffusion graphs as a method to rapidly generate vertex sequences for network embedding. Its computational efficiency is superior to previous methods due to simpler sequence generation, and it produces more accurate results. In experiments, we found that the performance relative to other methods improves with increasing edge density in the graph. In a community detection task, clustering nodes in the embedding space produces better results compared to other sequence based embedding methods. 
\end{abstract}

\input{./sections/intro.tex}
\input{./sections/related.tex}

\input{./sections/theory.tex}

\input{./sections/algorithm.tex}

\input{./sections/experiment.tex}
\input{./sections/conclusions.tex}

\end{document}

%% file: sections/intro.tex
%!TEX root = sequence.tex

\section{Introduction}

%Embedding of graphs -- utlities -- dimension reduction, visualisation, routing, distance oracles.  
%
%recent approaches sequence based embedding -- deeepwalk and node2vec (take time) due to cost of random walks
%
%
%we present a method... 

Embedding graphs into a low dimensional Euclidean spaces is a way of simplifying the graph information by associating each node with a point in the space. Thus, various methods of graph embedding have been developed and applied to different domains, such as visualisation~\cite{herman2000graph}, community and cluster identification~\cite{white2005spectral}, localisation of wireless devices~\cite{shang2003localization}, network routing~\cite{sarkar2009greedy} etc. Graph embeddings usually aim to preserve proximity -- nearby nodes on the graph should have similar coordinates -- in addition to properties specific to the application. 

In recent years, sequence based graph embedding methods have been developed as a way of generating Euclidean representations using sequence of vertices obtained from random walks. These methods are inspired by Word2Vec -- a method to embed words into Euclidean space based on sequences in which they occur. Word2vec takes short sequences of words from a document and uses them to train a neural network; in the process it obtains an embedding for the words. The embedding  space acts as an abstract latent space of {\em features}, and places two words close if they frequently occur nearby in the sequences \cite{mikolov2013efficient}. Sequence based graph embedding methods on the other hand obtain their vertex sequences by random walk on graphs and then apply analogous neural network methods for the embedding. The random walk has the advantage that it obtains a view of the neighborhood, without having to compute and store complete neighborhoods, which can be expensive in a large graph with many high degree vertices.

However, random walks are inefficient for generating proximity statistics. They are known to spread slowly, and revisit a vertex many times producing redundant information~\cite{alon2011many}. As a result, they require many steps or many restarts to cover the neighborhood of a node. Methods like Node2vec~\cite{grover2016node2vec} try to bias the walks away from recently visited nodes, but in the process they incur a cost due to the complexity of modifying transition probabilities with each step. We instead use a diffusion process that samples a subgraph of the neighborhood, from which several walks can be generated more efficiently. 

\vspace{-2em}
\input{./figures/kernel.tex}

\myparagraph{Our Contributions.} In our method, we extract a subgraph of the neighborhood of a node using a diffusion-like process, and call it a {\em diffusion graph}. On this subgraph, we compute an Euler tour to use as a sequence. By covering all adjacencies in the graph, the Euler tour contains a more complete view of the local neighborhood than random walks. We refer to this sequence generating method as {\em Diff2Vec} (D2V). The sequences generated by Diff2Vec are then used to train a neural network with one hidden layer containing $d$ neurons. The input weights of the neurons determine the embedding of the nodes.

Due to its better coverage of neighborhoods, Diff2Vec can operate with smaller neighborhood samples. As a result, it is more efficient than existing methods. In our experiments with a basic implementation, it turned out to be several times faster. In particular, it scales better with increasing density (vertex degrees) of graphs. Our experiments also show that the embedding preserves graph distances to a high accuracy. On experiments of community detection, we found that clustering applied to the embedding produces communities of high quality --  verified by the high modularity of the clusters. We publish a high performance parallel implementation of Diff2Vec which is available at \url{https://github.com/benedekrozemberczki/diff2vec}.

%The remainder of the paper is structured as follows. In Section \ref{sec:related} we overview related literature. The theoretical background on sequence based graph embedding is discussed in Section \ref{sec:theory}. Our sampling procedure is introduced in Section \ref{sec:algorithm}. Experimental results are analyzed in Section \ref{sec:experiments} and the paper concludes with Section \ref{sec:conclusions}.

%% file: figures/kernel.tex
\begin{figure}[h!]
\centering
	\subfloat[\label{fig:approxcdfs}]{
		\begin{tikzpicture}
		\begin{axis}[
		width=0.5\textwidth,
		height=0.4\textwidth,
		grid=major,
		grid style={dashed, gray!30},
		inner axis line style={-stealth},
		legend columns=3, 
		xlabel style={align=center},
		x label style={at={(axis description cs:0.5,1.4)},anchor=north},
		xlabel={\footnotesize Relative graph distance\\ approximation error},ylabel={\small CDF},
		legend style={
			at={(0.5,-0.45)},
			anchor=south}
		]
		\addplot[green,smooth,opacity=1, very thick] coordinates {
(0.0010773636253884,0.000765313696255315)
(0.120475872658999,0.150008331411617)
(0.272029108943386,0.301697109536262)
(0.423582345227773,0.458094432187706)
(0.575135581512161,0.620139534058732)
(0.726688817796548,0.788557837593154)
(0.878242054080935,0.972556947930842)
(1.00,0.981946361529151)

		};

		\addplot[red,smooth,opacity=1, very thick] coordinates {

(0.0007823593450121,0.000476300699617354)
(0.0250151156484477,0.185722827157957)
(0.0708125906419075,0.330806787567205)
(0.116610065635367,0.423317127161379)
(0.162407540628827,0.487280590750388)
(0.208205015622287,0.546257474601547)
(0.254002490615747,0.599540751204283)
(0.299799965609207,0.651688092136351)
(0.345597440602666,0.711505254212627)
(0.391394915596126,0.779040549743886)
(0.437192390589586,0.836309679585193)
(0.482989865583046,0.886474874900285)
(0.528787340576506,0.928855895208953)
(0.574584815569966,0.95919036250345)
(0.620382290563425,0.976746094692485)
(0.666179765556885,0.986815765366345)
(0.711977240550345,0.992819705245968)
(0.757774715543805,0.994701207190601)
(0.803572190537265,0.995168009908004)
(0.849369665530724,0.995244629080618)
(0.895167140524184,0.995296854835876)
(0.940964615517644,0.99535873361475)
(0.986762090511104,0.995443309937947)
(1.00,0.995525515157755)
			
		};
		
		\addplot[blue,smooth,opacity=1, very thick] coordinates {			

(0.00834567849138722,0.000485569779362712)
(0.0122328047431309,0.152368610016106)
(0.032811287977649,0.287208738395885)
(0.0533897712121671,0.398843332443732)
(0.0739682544466852,0.490300864539651)
(0.0945467376812033,0.568925160243119)
(0.115125220915721,0.639611107703479)
(0.13570370415024,0.705251074678969)
(0.156282187384758,0.76828119565858)
(0.176860670619276,0.827489070727695)
(0.197439153853794,0.877820027267087)
(0.218017637088312,0.916255192083285)
(0.23859612032283,0.943806786839008)
(0.259174603557348,0.962604841697519)
(0.279753086791866,0.975150395171007)
(0.300331570026384,0.983779725305042)
(0.320910053260902,0.990035785368152)
(0.341488536495421,0.993513667183732)
(0.362067019729939,0.995149068403586)
(0.382645502964457,0.995830099885609)
(0.403223986198975,0.9962073519268)
(0.423802469433493,0.99650331928968)
(0.444380952668011,0.996804028052181)
(0.464959435902529,0.997124044514712)
(0.485537919137047,0.997434339917574)
(0.506116402371566,0.997769921636519)
(0.526694885606084,0.998079424599429)
(0.547273368840602,0.998399850018797)
(0.56785185207512,0.998675542318647)
(0.588430335309638,0.998943283196358)
(0.609008818544156,0.999184410294543)
(0.629587301778674,0.999377395911335)
(0.650165785013192,0.999538011784598)
(0.67074426824771,0.999671392319537)
(0.691322751482228,0.999777101112945)
(0.711901234716747,0.999854152938041)
(0.732479717951265,0.999911379349755)
(0.753058201185783,0.999946584187795)
(0.773636684420301,0.999972133515271)
(0.794215167654819,0.999986275026546)
(0.814793650889337,0.999993321083548)
(0.835372134123855,0.999996727265834)
(0.855950617358373,0.99999800664575)
(0.876529100592891,0.999998921993405)
(0.89710758382741,0.999999979971696)
(0.917686067061928,0.99999998041093)
(0.938264550296446,0.999999989845718)
(0.958843033530964,0.999999989845718)
(0.979421516765482,0.999999989968945)
(1,1)
	
		};

		\addplot[green,smooth,opacity=1, very thick,dashed] coordinates {
(0.027078677540518,0.000125061076402925)
(0.040641284500499,0.0373359489053887)
(0.108361246541516,0.0762018852052696)
(0.176081208582533,0.117988241271344)
(0.24380117062355,0.163538954750728)
(0.311521132664567,0.213827014666997)
(0.379241094705584,0.269893173655658)
(0.446961056746601,0.332192524036893)
(0.514681018787618,0.401054905266947)
(0.582400980828635,0.476478487035127)
(0.650120942869652,0.558605808707525)
(0.71784090491067,0.646569661084856)
(0.785560866951687,0.740227682549116)
(0.853280828992704,0.838967422731794)
(0.921000791033721,0.946376755153666)
(0.988720753074738,0.99949175475107)
(1.00,0.99963510201644)

		};
				\addplot[red,smooth,opacity=1, very thick,dashed] coordinates {
					
							(0.0033760365497731,4.75403640520837e-06)
							(0.00832884433628935,0.0022392715515092)
							(0.0300337252223518,0.0053574069267038)
							(0.0517386061084143,0.0103068718279166)
							(0.0734434869944768,0.0181565411359025)
							(0.0951483678805392,0.0298702261994112)
							(0.116853248766602,0.0454327889388953)
							(0.138558129652664,0.0640598069562843)
							(0.160263010538727,0.0840807165196087)
							(0.181967891424789,0.103424160541601)
							(0.203672772310852,0.120404144193162)
							(0.225377653196914,0.134367301973174)
							(0.247082534082977,0.145586677227938)
							(0.268787414969039,0.154900395866709)
							(0.290492295855101,0.164317112803246)
							(0.312197176741164,0.177315955168783)
							(0.333902057627226,0.199850798039891)
							(0.355606938513289,0.238920353976968)
							(0.377311819399351,0.298301918134455)
							(0.399016700285414,0.372489915529585)
							(0.420721581171476,0.44957567048043)
							(0.442426462057539,0.518793756593326)
							(0.464131342943601,0.577046285332628)
							(0.485836223829664,0.6277899354818)
							(0.507541104715726,0.676376071103045)
							(0.529245985601789,0.725053386260974)
							(0.550950866487851,0.772455103410574)
							(0.572655747373914,0.816627524017732)
							(0.594360628259976,0.857859958317788)
							(0.616065509146038,0.895497517340259)
							(0.637770390032101,0.926615875857674)
							(0.659475270918163,0.949638288749543)
							(0.681180151804226,0.966256451011877)
							(0.702885032690288,0.978026289166228)
							(0.724589913576351,0.986359907575511)
							(0.746294794462413,0.992332453435382)
							(0.767999675348476,0.996382907333175)
							(0.789704556234538,0.998691311280225)
							(0.811409437120601,0.999624187919386)
							(0.833114318006663,0.999920509813145)
							(0.854819198892726,0.999985934799607)
							(0.876524079778788,0.99999431323464)
							(0.898228960664851,0.999997019301087)
							(0.919933841550913,0.999998598197861)
							(0.941638722436975,0.999999156596852)
							(0.963343603323038,0.999999413239819)
							(0.9850484842091,0.999999831599413)
							(1.000,0.999999942676557)
					};
	\addplot[blue,smooth,opacity=1, very thick,dashed] coordinates {
(0.00150203084566601,3.31182823410986e-05)
(0.0100014308208245,0.0170027791976299)
(0.0350231700983092,0.0345914193498618)
(0.0600449093757938,0.0523438931490265)
(0.0850666486532785,0.0705714297691003)
(0.110088387930763,0.0896912424429763)
(0.135110127208248,0.110264517089003)
(0.160131866485732,0.133991021507337)
(0.185153605763217,0.162359047505758)
(0.210175345040702,0.197515493675674)
(0.235197084318186,0.240740367270139)
(0.260218823595671,0.292246918120344)
(0.285240562873156,0.351181372624768)
(0.31026230215064,0.415189780664923)
(0.335284041428125,0.481783362012821)
(0.36030578070561,0.548144119130805)
(0.385327519983094,0.612442922334845)
(0.410349259260579,0.67334929788474)
(0.435370998538064,0.73010943940093)
(0.460392737815548,0.782168921137394)
(0.485414477093033,0.829535332248424)
(0.510436216370518,0.871578396786562)
(0.535457955648002,0.907715149026809)
(0.560479694925487,0.936548348215277)
(0.585501434202972,0.958246057626116)
(0.610523173480456,0.973449998443943)
(0.635544912757941,0.983742149749295)
(0.660566652035426,0.990513243043672)
(0.68558839131291,0.994778819347304)
(0.710610130590395,0.997370969398793)
(0.73563186986788,0.99879581738668)
(0.760653609145364,0.999465040544806)
(0.785675348422849,0.999764689046915)
(0.810697087700333,0.999900762077841)
(0.835718826977818,0.999958033494824)
(0.860740566255303,0.999975911858854)
(0.885762305532787,0.999985210005359)
(0.910784044810272,0.999990052525483)
(0.935805784087757,0.99999318094772)
(0.960827523365241,0.999995257433616)
(0.985849262642726,0.999997012531274)
(1.000,0.999997976972135)		
		
									};
		\legend{\scriptsize\text{D2V}  \textit{d}=2, \textit{d}=32, \textit{d}=128,\scriptsize\text{N2V}  \textit{d}=2, \textit{d}=32,  \textit{d}=128}
		\end{axis}		
		\end{tikzpicture}}
\hfill
\subfloat[\label{fig:watts}]{\begin{tikzpicture}[scale=0.29,transform shape]
	\tikzstyle{VertexStyle}=[minimum size = 10pt,inner sep=0pt, shape = circle]
	\Vertex[L=$$,x=0.0,y=7.485]{0}
	\Vertex[L=$$,x=0.48,y=7.335]{1}
	\Vertex[L=$$,x=0.78,y=6.96]{2}
	\Vertex[L=$$,x=0.435,y=6.705]{3}
	\Vertex[L=$$,x=0.63,y=6.45]{4}
	\Vertex[L=$$,x=0.615,y=6.105]{5}
	\Vertex[L=$$,x=0.81,y=5.85]{6}
	\Vertex[L=$$,x=0.78,y=5.46]{7}
	\Vertex[L=$$,x=0.495,y=5.145]{8}
	\Vertex[L=$$,x=0.9,y=4.995]{9}
	\Vertex[L=$$,x=1.02,y=4.635]{10}
	\Vertex[L=$$,x=1.26,y=4.41]{11}
	\Vertex[L=$$,x=1.575,y=4.155]{12}
	\Vertex[L=$$,x=1.83,y=3.945]{13}
	\Vertex[L=$$,x=9.24,y=6.84]{14}
	\Vertex[L=$$,x=1.89,y=3.33]{15}
	\Vertex[L=$$,x=2.235,y=3.165]{16}
	\Vertex[L=$$,x=2.325,y=2.835]{17}
	\Vertex[L=$$,x=2.895,y=2.805]{18}
	\Vertex[L=$$,x=3.21,y=2.805]{19}
	\Vertex[L=$$,x=3.3,y=2.55]{20}
	\Vertex[L=$$,x=3.615,y=2.43]{21}
	\Vertex[L=$$,x=4.02,y=2.34]{22}
	\Vertex[L=$$,x=4.245,y=2.115]{23}
	\Vertex[L=$$,x=9.06,y=6.54]{24}
	\Vertex[L=$$,x=4.59,y=1.755]{25}
	\Vertex[L=$$,x=4.725,y=1.56]{26}
	\Vertex[L=$$,x=4.995,y=1.41]{27}
	\Vertex[L=$$,x=4.995,y=0.735]{28}
	\Vertex[L=$$,x=5.28,y=0.54]{29}
	\Vertex[L=$$,x=8.805,y=5.805]{30}
	\Vertex[L=$$,x=6.03,y=0.885]{31}
	\Vertex[L=$$,x=6.135,y=0.465]{32}
	\Vertex[L=$$,x=6.48,y=0.36]{33}
	\Vertex[L=$$,x=6.78,y=0.06]{34}
	\Vertex[L=$$,x=7.125,y=0.42]{35}
	\Vertex[L=$$,x=7.365,y=0.255]{36}
	\Vertex[L=$$,x=7.65,y=0.195]{37}
	\Vertex[L=$$,x=7.8,y=0.105]{38}
	\Vertex[L=$$,x=8.265,y=0.21]{39}
	\Vertex[L=$$,x=8.49,y=0.03]{40}
	\Vertex[L=$$,x=8.835,y=0.135]{41}
	\Vertex[L=$$,x=9.21,y=0.0]{42}
	\Vertex[L=$$,x=9.375,y=0.18]{43}
	\Vertex[L=$$,x=9.69,y=0.15]{44}
	\Vertex[L=$$,x=9.96,y=0.375]{45}
	\Vertex[L=$$,x=10.26,y=0.51]{46}
	\Vertex[L=$$,x=10.53,y=0.63]{47}
	\Vertex[L=$$,x=10.785,y=0.6]{48}
	\Vertex[L=$$,x=11.025,y=0.945]{49}
	\Vertex[L=$$,x=11.34,y=1.065]{50}
	\Vertex[L=$$,x=11.55,y=1.275]{51}
	\Vertex[L=$$,x=11.82,y=1.41]{52}
	\Vertex[L=$$,x=11.985,y=1.65]{53}
	\Vertex[L=$$,x=12.24,y=1.89]{54}
	\Vertex[L=$$,x=12.405,y=2.22]{55}
	\Vertex[L=$$,x=12.705,y=2.31]{56}
	\Vertex[L=$$,x=12.825,y=2.61]{57}
	\Vertex[L=$$,x=12.975,y=3.0]{58}
	\Vertex[L=$$,x=13.185,y=3.165]{59}
	\Vertex[L=$$,x=13.275,y=3.48]{60}
	\Vertex[L=$$,x=13.545,y=3.735]{61}
	\Vertex[L=$$,x=12.12,y=5.205]{62}
	\Vertex[L=$$,x=13.815,y=4.23]{63}
	\Vertex[L=$$,x=14.055,y=4.455]{64}
	\Vertex[L=$$,x=14.085,y=4.77]{65}
	\Vertex[L=$$,x=14.22,y=5.055]{66}
	\Vertex[L=$$,x=14.28,y=5.34]{67}
	\Vertex[L=$$,x=14.505,y=5.58]{68}
	\Vertex[L=$$,x=14.58,y=5.88]{69}
	\Vertex[L=$$,x=14.7,y=6.21]{70}
	\Vertex[L=$$,x=14.7,y=6.405]{71}
	\Vertex[L=$$,x=15.0,y=6.765]{72}
	\Vertex[L=$$,x=14.88,y=7.125]{73}
	\Vertex[L=$$,x=15.0,y=7.395]{74}
	\Vertex[L=$$,x=14.895,y=7.695]{75}
	\Vertex[L=$$,x=14.79,y=7.905]{76}
	\Vertex[L=$$,x=14.94,y=8.37]{77}
	\Vertex[L=$$,x=14.595,y=8.565]{78}
	\Vertex[L=$$,x=14.685,y=8.865]{79}
	\Vertex[L=$$,x=14.805,y=9.225]{80}
	\Vertex[L=$$,x=14.67,y=9.45]{81}
	\Vertex[L=$$,x=14.58,y=9.795]{82}
	\Vertex[L=$$,x=10.77,y=7.965]{83}
	\Vertex[L=$$,x=14.385,y=10.395]{84}
	\Vertex[L=$$,x=14.235,y=10.575]{85}
	\Vertex[L=$$,x=14.1,y=10.86]{86}
	\Vertex[L=$$,x=13.92,y=11.13]{87}
	\Vertex[L=$$,x=13.995,y=11.595]{88}
	\Vertex[L=$$,x=13.875,y=11.745]{89}
	\Vertex[L=$$,x=13.665,y=12.015]{90}
	\Vertex[L=$$,x=13.515,y=12.195]{91}
	\Vertex[L=$$,x=13.29,y=12.42]{92}
	\Vertex[L=$$,x=13.08,y=12.645]{93}
	\Vertex[L=$$,x=12.915,y=12.855]{94}
	\Vertex[L=$$,x=12.69,y=12.87]{95}
	\Vertex[L=$$,x=12.63,y=13.635]{96}
	\Vertex[L=$$,x=12.405,y=13.815]{97}
	\Vertex[L=$$,x=11.985,y=13.68]{98}
	\Vertex[L=$$,x=11.79,y=13.845]{99}
	\Vertex[L=$$,x=11.505,y=13.965]{100}
	\Vertex[L=$$,x=11.295,y=14.19]{101}
	\Vertex[L=$$,x=11.07,y=14.265]{102}
	\Vertex[L=$$,x=10.26,y=11.355]{103}
	\Vertex[L=$$,x=10.545,y=14.79]{104}
	\Vertex[L=$$,x=10.32,y=14.685]{105}
	\Vertex[L=$$,x=9.945,y=14.58]{106}
	\Vertex[L=$$,x=9.975,y=13.14]{107}
	\Vertex[L=$$,x=9.405,y=14.76]{108}
	\Vertex[L=$$,x=9.21,y=14.73]{109}
	\Vertex[L=$$,x=9.27,y=12.27]{110}
	\Vertex[L=$$,x=8.475,y=14.715]{111}
	\Vertex[L=$$,x=8.355,y=14.595]{112}
	\Vertex[L=$$,x=7.935,y=15.0]{113}
	\Vertex[L=$$,x=8.85,y=10.575]{114}
	\Vertex[L=$$,x=7.305,y=14.76]{115}
	\Vertex[L=$$,x=6.96,y=14.88]{116}
	\Vertex[L=$$,x=6.735,y=14.685]{117}
	\Vertex[L=$$,x=6.42,y=14.58]{118}
	\Vertex[L=$$,x=6.225,y=14.595]{119}
	\Vertex[L=$$,x=5.76,y=14.73]{120}
	\Vertex[L=$$,x=5.49,y=14.49]{121}
	\Vertex[L=$$,x=7.29,y=11.475]{122}
	\Vertex[L=$$,x=4.815,y=14.355]{123}
	\Vertex[L=$$,x=4.8,y=14.1]{124}
	\Vertex[L=$$,x=4.62,y=13.8]{125}
	\Vertex[L=$$,x=4.245,y=13.74]{126}
	\Vertex[L=$$,x=4.02,y=13.53]{127}
	\Vertex[L=$$,x=3.63,y=13.47]{128}
	\Vertex[L=$$,x=3.495,y=13.29]{129}
	\Vertex[L=$$,x=3.375,y=12.87]{130}
	\Vertex[L=$$,x=3.06,y=12.81]{131}
	\Vertex[L=$$,x=2.7,y=12.69]{132}
	\Vertex[L=$$,x=4.155,y=11.205]{133}
	\Vertex[L=$$,x=2.28,y=12.27]{134}
	\Vertex[L=$$,x=2.205,y=11.94]{135}
	\Vertex[L=$$,x=1.77,y=11.85]{136}
	\Vertex[L=$$,x=1.74,y=11.505]{137}
	\Vertex[L=$$,x=1.395,y=11.385]{138}
	\Vertex[L=$$,x=1.575,y=10.92]{139}
	\Vertex[L=$$,x=1.5,y=10.65]{140}
	\Vertex[L=$$,x=1.065,y=10.41]{141}
	\Vertex[L=$$,x=1.23,y=10.035]{142}
	\Vertex[L=$$,x=0.96,y=9.93]{143}
	\Vertex[L=$$,x=0.975,y=9.45]{144}
	\Vertex[L=$$,x=0.54,y=9.42]{145}
	\Vertex[L=$$,x=0.885,y=8.835]{146}
	\Vertex[L=$$,x=2.475,y=8.22]{147}
	\Vertex[L=$$,x=4.44,y=7.59]{148}
	\Vertex[L=$$,x=0.9,y=7.95]{149}
	\AddVertexColor{gray!90, draw=black}{0,1,2,3,4,5,6,7,8,9,10,11,12,13,14,15,16,17,18,19,20,21,22,23,24,25,26,27,28,29,30,31,32,33,34,35,36,37,38,39,40,41,42,43,44,45,46,47,48,49,50,51,52,53,54,55,56,57,58,59,60,61,62,63,64,65,66,67,68,69,70,71,72,73,74,75,76,77,78,79,80,81,82,83,84,85,86,87,88,89,90,91,92,93,94,95,96,97,98,99,100,101,102,103,104,105,106,107,108,109,110,111,112,113,114,115,116,117,118,119,120,121,122,123,124,125,126,127,128,129,130,131,132,133,134,135,136,137,138,139,140,141,142,143,144,145,146,147,148,149}
	\tikzstyle{EdgeStyle}=[line width=0.7pt, opacity = 0.3]
	\tikzstyle{LabelStyle}=[fill=white]
	\Edge[label=](0)(1)
	\Edge[label=](0)(2)
	\Edge[label=](0)(3)
	\Edge[label=](0)(4)
	\Edge[label=](0)(5)
	\Edge[label=](0)(107)
	\Edge[label=](0)(145)
	\Edge[label=](0)(146)
	\Edge[label=](0)(147)
	\Edge[label=](0)(148)
	\Edge[label=](0)(149)
	\Edge[label=](1)(2)
	\Edge[label=](1)(3)
	\Edge[label=](1)(4)
	\Edge[label=](1)(5)
	\Edge[label=](1)(6)
	\Edge[label=](1)(146)
	\Edge[label=](1)(147)
	\Edge[label=](1)(148)
	\Edge[label=](1)(149)
	\Edge[label=](2)(3)
	\Edge[label=](2)(4)
	\Edge[label=](2)(5)
	\Edge[label=](2)(6)
	\Edge[label=](2)(7)
	\Edge[label=](2)(147)
	\Edge[label=](2)(148)
	\Edge[label=](2)(149)
	\Edge[label=](3)(4)
	\Edge[label=](3)(5)
	\Edge[label=](3)(6)
	\Edge[label=](3)(7)
	\Edge[label=](3)(8)
	\Edge[label=](3)(148)
	\Edge[label=](3)(149)
	\Edge[label=](3)(62)
	\Edge[label=](4)(5)
	\Edge[label=](4)(6)
	\Edge[label=](4)(7)
	\Edge[label=](4)(8)
	\Edge[label=](4)(81)
	\Edge[label=](4)(149)
	\Edge[label=](5)(6)
	\Edge[label=](5)(8)
	\Edge[label=](5)(9)
	\Edge[label=](5)(10)
	\Edge[label=](5)(111)
	\Edge[label=](6)(7)
	\Edge[label=](6)(8)
	\Edge[label=](6)(9)
	\Edge[label=](6)(10)
	\Edge[label=](6)(11)
	\Edge[label=](6)(37)
	\Edge[label=](7)(8)
	\Edge[label=](7)(9)
	\Edge[label=](7)(11)
	\Edge[label=](7)(12)
	\Edge[label=](7)(93)
	\Edge[label=](8)(9)
	\Edge[label=](8)(10)
	\Edge[label=](8)(11)
	\Edge[label=](8)(12)
	\Edge[label=](8)(13)
	\Edge[label=](9)(10)
	\Edge[label=](9)(11)
	\Edge[label=](9)(12)
	\Edge[label=](9)(13)
	\Edge[label=](9)(14)
	\Edge[label=](10)(11)
	\Edge[label=](10)(12)
	\Edge[label=](10)(13)
	\Edge[label=](10)(14)
	\Edge[label=](10)(15)
	\Edge[label=](11)(12)
	\Edge[label=](11)(13)
	\Edge[label=](11)(14)
	\Edge[label=](11)(15)
	\Edge[label=](11)(16)
	\Edge[label=](12)(13)
	\Edge[label=](12)(14)
	\Edge[label=](12)(15)
	\Edge[label=](12)(16)
	\Edge[label=](12)(17)
	\Edge[label=](13)(14)
	\Edge[label=](13)(15)
	\Edge[label=](13)(16)
	\Edge[label=](13)(17)
	\Edge[label=](13)(18)
	\Edge[label=](14)(15)
	\Edge[label=](14)(16)
	\Edge[label=](14)(17)
	\Edge[label=](14)(18)
	\Edge[label=](14)(19)
	\Edge[label=](15)(16)
	\Edge[label=](15)(17)
	\Edge[label=](15)(18)
	\Edge[label=](15)(19)
	\Edge[label=](15)(20)
	\Edge[label=](16)(17)
	\Edge[label=](16)(18)
	\Edge[label=](16)(19)
	\Edge[label=](16)(20)
	\Edge[label=](16)(21)
	\Edge[label=](17)(18)
	\Edge[label=](17)(19)
	\Edge[label=](17)(20)
	\Edge[label=](17)(21)
	\Edge[label=](17)(22)
	\Edge[label=](18)(19)
	\Edge[label=](18)(20)
	\Edge[label=](18)(21)
	\Edge[label=](18)(22)
	\Edge[label=](18)(23)
	\Edge[label=](19)(20)
	\Edge[label=](19)(21)
	\Edge[label=](19)(22)
	\Edge[label=](19)(23)
	\Edge[label=](19)(24)
	\Edge[label=](19)(143)
	\Edge[label=](20)(21)
	\Edge[label=](20)(22)
	\Edge[label=](20)(23)
	\Edge[label=](20)(24)
	\Edge[label=](20)(25)
	\Edge[label=](21)(83)
	\Edge[label=](21)(100)
	\Edge[label=](21)(22)
	\Edge[label=](21)(23)
	\Edge[label=](21)(24)
	\Edge[label=](21)(25)
	\Edge[label=](21)(26)
	\Edge[label=](22)(77)
	\Edge[label=](22)(23)
	\Edge[label=](22)(24)
	\Edge[label=](22)(26)
	\Edge[label=](22)(27)
	\Edge[label=](23)(24)
	\Edge[label=](23)(25)
	\Edge[label=](23)(26)
	\Edge[label=](23)(27)
	\Edge[label=](23)(28)
	\Edge[label=](24)(128)
	\Edge[label=](24)(26)
	\Edge[label=](24)(27)
	\Edge[label=](24)(28)
	\Edge[label=](24)(29)
	\Edge[label=](25)(107)
	\Edge[label=](25)(43)
	\Edge[label=](25)(114)
	\Edge[label=](25)(26)
	\Edge[label=](25)(27)
	\Edge[label=](25)(28)
	\Edge[label=](25)(30)
	\Edge[label=](26)(27)
	\Edge[label=](26)(28)
	\Edge[label=](26)(29)
	\Edge[label=](26)(30)
	\Edge[label=](26)(31)
	\Edge[label=](27)(32)
	\Edge[label=](27)(28)
	\Edge[label=](27)(29)
	\Edge[label=](27)(30)
	\Edge[label=](27)(31)
	\Edge[label=](28)(32)
	\Edge[label=](28)(33)
	\Edge[label=](28)(29)
	\Edge[label=](28)(30)
	\Edge[label=](28)(31)
	\Edge[label=](29)(32)
	\Edge[label=](29)(33)
	\Edge[label=](29)(34)
	\Edge[label=](29)(30)
	\Edge[label=](29)(31)
	\Edge[label=](30)(32)
	\Edge[label=](30)(33)
	\Edge[label=](30)(34)
	\Edge[label=](30)(35)
	\Edge[label=](30)(31)
	\Edge[label=](31)(32)
	\Edge[label=](31)(33)
	\Edge[label=](31)(34)
	\Edge[label=](31)(35)
	\Edge[label=](31)(36)
	\Edge[label=](32)(33)
	\Edge[label=](32)(34)
	\Edge[label=](32)(35)
	\Edge[label=](32)(36)
	\Edge[label=](32)(37)
	\Edge[label=](33)(34)
	\Edge[label=](33)(35)
	\Edge[label=](33)(36)
	\Edge[label=](33)(37)
	\Edge[label=](33)(38)
	\Edge[label=](34)(35)
	\Edge[label=](34)(36)
	\Edge[label=](34)(37)
	\Edge[label=](34)(38)
	\Edge[label=](34)(39)
	\Edge[label=](35)(36)
	\Edge[label=](35)(37)
	\Edge[label=](35)(38)
	\Edge[label=](35)(39)
	\Edge[label=](35)(40)
	\Edge[label=](35)(104)
	\Edge[label=](36)(37)
	\Edge[label=](36)(38)
	\Edge[label=](36)(39)
	\Edge[label=](36)(40)
	\Edge[label=](36)(41)
	\Edge[label=](37)(38)
	\Edge[label=](37)(39)
	\Edge[label=](37)(40)
	\Edge[label=](37)(41)
	\Edge[label=](38)(39)
	\Edge[label=](38)(40)
	\Edge[label=](38)(41)
	\Edge[label=](38)(42)
	\Edge[label=](38)(43)
	\Edge[label=](39)(40)
	\Edge[label=](39)(41)
	\Edge[label=](39)(42)
	\Edge[label=](39)(43)
	\Edge[label=](39)(44)
	\Edge[label=](40)(41)
	\Edge[label=](40)(42)
	\Edge[label=](40)(43)
	\Edge[label=](40)(45)
	\Edge[label=](40)(69)
	\Edge[label=](41)(42)
	\Edge[label=](41)(43)
	\Edge[label=](41)(44)
	\Edge[label=](41)(45)
	\Edge[label=](41)(46)
	\Edge[label=](42)(43)
	\Edge[label=](42)(44)
	\Edge[label=](42)(45)
	\Edge[label=](42)(46)
	\Edge[label=](42)(47)
	\Edge[label=](43)(44)
	\Edge[label=](43)(46)
	\Edge[label=](43)(47)
	\Edge[label=](43)(48)
	\Edge[label=](44)(45)
	\Edge[label=](44)(46)
	\Edge[label=](44)(47)
	\Edge[label=](44)(48)
	\Edge[label=](44)(49)
	\Edge[label=](45)(46)
	\Edge[label=](45)(47)
	\Edge[label=](45)(48)
	\Edge[label=](45)(49)
	\Edge[label=](45)(50)
	\Edge[label=](46)(47)
	\Edge[label=](46)(48)
	\Edge[label=](46)(49)
	\Edge[label=](46)(50)
	\Edge[label=](46)(51)
	\Edge[label=](47)(48)
	\Edge[label=](47)(49)
	\Edge[label=](47)(50)
	\Edge[label=](47)(51)
	\Edge[label=](47)(52)
	\Edge[label=](48)(49)
	\Edge[label=](48)(50)
	\Edge[label=](48)(51)
	\Edge[label=](48)(52)
	\Edge[label=](48)(53)
	\Edge[label=](49)(50)
	\Edge[label=](49)(51)
	\Edge[label=](49)(52)
	\Edge[label=](49)(53)
	\Edge[label=](49)(54)
	\Edge[label=](50)(52)
	\Edge[label=](50)(54)
	\Edge[label=](50)(55)
	\Edge[label=](50)(84)
	\Edge[label=](50)(143)
	\Edge[label=](51)(52)
	\Edge[label=](51)(53)
	\Edge[label=](51)(54)
	\Edge[label=](51)(55)
	\Edge[label=](51)(56)
	\Edge[label=](52)(96)
	\Edge[label=](52)(53)
	\Edge[label=](52)(54)
	\Edge[label=](52)(55)
	\Edge[label=](52)(56)
	\Edge[label=](52)(57)
	\Edge[label=](53)(128)
	\Edge[label=](53)(54)
	\Edge[label=](53)(55)
	\Edge[label=](53)(56)
	\Edge[label=](53)(57)
	\Edge[label=](53)(58)
	\Edge[label=](54)(55)
	\Edge[label=](54)(56)
	\Edge[label=](54)(57)
	\Edge[label=](54)(58)
	\Edge[label=](54)(59)
	\Edge[label=](55)(56)
	\Edge[label=](55)(57)
	\Edge[label=](55)(58)
	\Edge[label=](55)(59)
	\Edge[label=](55)(60)
	\Edge[label=](55)(90)
	\Edge[label=](56)(145)
	\Edge[label=](56)(57)
	\Edge[label=](56)(58)
	\Edge[label=](56)(59)
	\Edge[label=](56)(60)
	\Edge[label=](56)(61)
	\Edge[label=](57)(58)
	\Edge[label=](57)(59)
	\Edge[label=](57)(60)
	\Edge[label=](57)(61)
	\Edge[label=](57)(62)
	\Edge[label=](58)(83)
	\Edge[label=](58)(60)
	\Edge[label=](58)(61)
	\Edge[label=](58)(62)
	\Edge[label=](58)(63)
	\Edge[label=](59)(64)
	\Edge[label=](59)(75)
	\Edge[label=](59)(60)
	\Edge[label=](59)(61)
	\Edge[label=](59)(63)
	\Edge[label=](60)(64)
	\Edge[label=](60)(65)
	\Edge[label=](60)(61)
	\Edge[label=](60)(62)
	\Edge[label=](60)(63)
	\Edge[label=](61)(64)
	\Edge[label=](61)(65)
	\Edge[label=](61)(66)
	\Edge[label=](61)(62)
	\Edge[label=](61)(63)
	\Edge[label=](62)(64)
	\Edge[label=](62)(65)
	\Edge[label=](62)(66)
	\Edge[label=](62)(63)
	\Edge[label=](63)(64)
	\Edge[label=](63)(65)
	\Edge[label=](63)(66)
	\Edge[label=](63)(67)
	\Edge[label=](63)(68)
	\Edge[label=](64)(65)
	\Edge[label=](64)(66)
	\Edge[label=](64)(67)
	\Edge[label=](64)(68)
	\Edge[label=](64)(69)
	\Edge[label=](64)(90)
	\Edge[label=](65)(66)
	\Edge[label=](65)(67)
	\Edge[label=](65)(68)
	\Edge[label=](65)(69)
	\Edge[label=](65)(70)
	\Edge[label=](66)(67)
	\Edge[label=](66)(68)
	\Edge[label=](66)(69)
	\Edge[label=](66)(70)
	\Edge[label=](66)(71)
	\Edge[label=](67)(68)
	\Edge[label=](67)(69)
	\Edge[label=](67)(70)
	\Edge[label=](67)(71)
	\Edge[label=](67)(72)
	\Edge[label=](68)(69)
	\Edge[label=](68)(70)
	\Edge[label=](68)(71)
	\Edge[label=](68)(72)
	\Edge[label=](68)(73)
	\Edge[label=](69)(70)
	\Edge[label=](69)(71)
	\Edge[label=](69)(72)
	\Edge[label=](69)(73)
	\Edge[label=](69)(74)
	\Edge[label=](70)(71)
	\Edge[label=](70)(72)
	\Edge[label=](70)(73)
	\Edge[label=](70)(74)
	\Edge[label=](70)(75)
	\Edge[label=](71)(72)
	\Edge[label=](71)(73)
	\Edge[label=](71)(74)
	\Edge[label=](71)(75)
	\Edge[label=](71)(76)
	\Edge[label=](72)(73)
	\Edge[label=](72)(74)
	\Edge[label=](72)(75)
	\Edge[label=](72)(76)
	\Edge[label=](72)(77)
	\Edge[label=](73)(74)
	\Edge[label=](73)(75)
	\Edge[label=](73)(76)
	\Edge[label=](73)(77)
	\Edge[label=](73)(78)
	\Edge[label=](74)(75)
	\Edge[label=](74)(76)
	\Edge[label=](74)(77)
	\Edge[label=](74)(78)
	\Edge[label=](74)(79)
	\Edge[label=](75)(76)
	\Edge[label=](75)(77)
	\Edge[label=](75)(78)
	\Edge[label=](75)(79)
	\Edge[label=](75)(80)
	\Edge[label=](76)(77)
	\Edge[label=](76)(78)
	\Edge[label=](76)(79)
	\Edge[label=](76)(80)
	\Edge[label=](76)(81)
	\Edge[label=](77)(78)
	\Edge[label=](77)(79)
	\Edge[label=](77)(80)
	\Edge[label=](77)(81)
	\Edge[label=](77)(82)
	\Edge[label=](78)(79)
	\Edge[label=](78)(80)
	\Edge[label=](78)(81)
	\Edge[label=](78)(82)
	\Edge[label=](78)(83)
	\Edge[label=](79)(80)
	\Edge[label=](79)(81)
	\Edge[label=](79)(82)
	\Edge[label=](79)(83)
	\Edge[label=](79)(84)
	\Edge[label=](80)(81)
	\Edge[label=](80)(82)
	\Edge[label=](80)(83)
	\Edge[label=](80)(84)
	\Edge[label=](80)(85)
	\Edge[label=](81)(82)
	\Edge[label=](81)(83)
	\Edge[label=](81)(84)
	\Edge[label=](81)(85)
	\Edge[label=](81)(86)
	\Edge[label=](82)(83)
	\Edge[label=](82)(84)
	\Edge[label=](82)(85)
	\Edge[label=](82)(86)
	\Edge[label=](82)(87)
	\Edge[label=](83)(85)
	\Edge[label=](83)(86)
	\Edge[label=](83)(87)
	\Edge[label=](83)(88)
	\Edge[label=](84)(85)
	\Edge[label=](84)(86)
	\Edge[label=](84)(87)
	\Edge[label=](84)(88)
	\Edge[label=](84)(89)
	\Edge[label=](85)(114)
	\Edge[label=](85)(87)
	\Edge[label=](85)(88)
	\Edge[label=](85)(89)
	\Edge[label=](85)(90)
	\Edge[label=](86)(87)
	\Edge[label=](86)(88)
	\Edge[label=](86)(89)
	\Edge[label=](86)(90)
	\Edge[label=](86)(91)
	\Edge[label=](87)(88)
	\Edge[label=](87)(89)
	\Edge[label=](87)(90)
	\Edge[label=](87)(91)
	\Edge[label=](87)(92)
	\Edge[label=](88)(89)
	\Edge[label=](88)(90)
	\Edge[label=](88)(91)
	\Edge[label=](88)(92)
	\Edge[label=](88)(93)
	\Edge[label=](89)(90)
	\Edge[label=](89)(91)
	\Edge[label=](89)(92)
	\Edge[label=](89)(93)
	\Edge[label=](89)(94)
	\Edge[label=](90)(113)
	\Edge[label=](90)(91)
	\Edge[label=](90)(92)
	\Edge[label=](90)(95)
	\Edge[label=](91)(96)
	\Edge[label=](91)(139)
	\Edge[label=](91)(92)
	\Edge[label=](91)(93)
	\Edge[label=](91)(94)
	\Edge[label=](92)(96)
	\Edge[label=](92)(97)
	\Edge[label=](92)(93)
	\Edge[label=](92)(94)
	\Edge[label=](92)(95)
	\Edge[label=](93)(96)
	\Edge[label=](93)(97)
	\Edge[label=](93)(98)
	\Edge[label=](93)(94)
	\Edge[label=](93)(95)
	\Edge[label=](94)(96)
	\Edge[label=](94)(97)
	\Edge[label=](94)(98)
	\Edge[label=](94)(99)
	\Edge[label=](94)(95)
	\Edge[label=](95)(96)
	\Edge[label=](95)(97)
	\Edge[label=](95)(98)
	\Edge[label=](95)(99)
	\Edge[label=](95)(100)
	\Edge[label=](96)(97)
	\Edge[label=](96)(98)
	\Edge[label=](96)(99)
	\Edge[label=](96)(100)
	\Edge[label=](97)(98)
	\Edge[label=](97)(99)
	\Edge[label=](97)(100)
	\Edge[label=](97)(101)
	\Edge[label=](97)(102)
	\Edge[label=](98)(99)
	\Edge[label=](98)(100)
	\Edge[label=](98)(101)
	\Edge[label=](98)(102)
	\Edge[label=](98)(103)
	\Edge[label=](99)(100)
	\Edge[label=](99)(101)
	\Edge[label=](99)(102)
	\Edge[label=](99)(103)
	\Edge[label=](99)(104)
	\Edge[label=](100)(101)
	\Edge[label=](100)(102)
	\Edge[label=](100)(103)
	\Edge[label=](100)(104)
	\Edge[label=](101)(102)
	\Edge[label=](101)(103)
	\Edge[label=](101)(104)
	\Edge[label=](101)(105)
	\Edge[label=](101)(106)
	\Edge[label=](102)(103)
	\Edge[label=](102)(104)
	\Edge[label=](102)(105)
	\Edge[label=](102)(106)
	\Edge[label=](102)(107)
	\Edge[label=](103)(104)
	\Edge[label=](103)(105)
	\Edge[label=](103)(106)
	\Edge[label=](103)(107)
	\Edge[label=](103)(108)
	\Edge[label=](104)(105)
	\Edge[label=](104)(106)
	\Edge[label=](104)(107)
	\Edge[label=](104)(109)
	\Edge[label=](105)(106)
	\Edge[label=](105)(107)
	\Edge[label=](105)(108)
	\Edge[label=](105)(109)
	\Edge[label=](105)(110)
	\Edge[label=](106)(107)
	\Edge[label=](106)(108)
	\Edge[label=](106)(109)
	\Edge[label=](106)(110)
	\Edge[label=](106)(111)
	\Edge[label=](107)(108)
	\Edge[label=](107)(109)
	\Edge[label=](107)(111)
	\Edge[label=](108)(109)
	\Edge[label=](108)(110)
	\Edge[label=](108)(111)
	\Edge[label=](108)(112)
	\Edge[label=](108)(113)
	\Edge[label=](109)(110)
	\Edge[label=](109)(111)
	\Edge[label=](109)(112)
	\Edge[label=](109)(113)
	\Edge[label=](109)(114)
	\Edge[label=](110)(111)
	\Edge[label=](110)(112)
	\Edge[label=](110)(113)
	\Edge[label=](110)(114)
	\Edge[label=](110)(115)
	\Edge[label=](111)(112)
	\Edge[label=](111)(113)
	\Edge[label=](111)(114)
	\Edge[label=](111)(115)
	\Edge[label=](111)(116)
	\Edge[label=](112)(113)
	\Edge[label=](112)(114)
	\Edge[label=](112)(115)
	\Edge[label=](112)(116)
	\Edge[label=](112)(117)
	\Edge[label=](113)(114)
	\Edge[label=](113)(115)
	\Edge[label=](113)(116)
	\Edge[label=](113)(118)
	\Edge[label=](114)(115)
	\Edge[label=](114)(116)
	\Edge[label=](114)(117)
	\Edge[label=](114)(118)
	\Edge[label=](114)(119)
	\Edge[label=](115)(116)
	\Edge[label=](115)(117)
	\Edge[label=](115)(118)
	\Edge[label=](115)(119)
	\Edge[label=](115)(120)
	\Edge[label=](116)(117)
	\Edge[label=](116)(118)
	\Edge[label=](116)(119)
	\Edge[label=](116)(120)
	\Edge[label=](116)(121)
	\Edge[label=](117)(118)
	\Edge[label=](117)(119)
	\Edge[label=](117)(120)
	\Edge[label=](117)(121)
	\Edge[label=](117)(122)
	\Edge[label=](118)(119)
	\Edge[label=](118)(120)
	\Edge[label=](118)(121)
	\Edge[label=](118)(122)
	\Edge[label=](118)(123)
	\Edge[label=](119)(139)
	\Edge[label=](119)(120)
	\Edge[label=](119)(122)
	\Edge[label=](119)(123)
	\Edge[label=](119)(124)
	\Edge[label=](120)(121)
	\Edge[label=](120)(122)
	\Edge[label=](120)(123)
	\Edge[label=](120)(124)
	\Edge[label=](120)(125)
	\Edge[label=](121)(122)
	\Edge[label=](121)(123)
	\Edge[label=](121)(124)
	\Edge[label=](121)(125)
	\Edge[label=](121)(126)
	\Edge[label=](122)(123)
	\Edge[label=](122)(124)
	\Edge[label=](122)(125)
	\Edge[label=](122)(126)
	\Edge[label=](122)(127)
	\Edge[label=](123)(128)
	\Edge[label=](123)(124)
	\Edge[label=](123)(125)
	\Edge[label=](123)(126)
	\Edge[label=](123)(127)
	\Edge[label=](124)(128)
	\Edge[label=](124)(129)
	\Edge[label=](124)(125)
	\Edge[label=](124)(126)
	\Edge[label=](124)(127)
	\Edge[label=](125)(128)
	\Edge[label=](125)(129)
	\Edge[label=](125)(130)
	\Edge[label=](125)(126)
	\Edge[label=](125)(127)
	\Edge[label=](126)(128)
	\Edge[label=](126)(129)
	\Edge[label=](126)(130)
	\Edge[label=](126)(131)
	\Edge[label=](126)(127)
	\Edge[label=](127)(128)
	\Edge[label=](127)(129)
	\Edge[label=](127)(130)
	\Edge[label=](127)(131)
	\Edge[label=](127)(132)
	\Edge[label=](128)(129)
	\Edge[label=](128)(131)
	\Edge[label=](128)(132)
	\Edge[label=](128)(133)
	\Edge[label=](129)(130)
	\Edge[label=](129)(131)
	\Edge[label=](129)(132)
	\Edge[label=](129)(134)
	\Edge[label=](129)(145)
	\Edge[label=](130)(131)
	\Edge[label=](130)(132)
	\Edge[label=](130)(133)
	\Edge[label=](130)(134)
	\Edge[label=](130)(135)
	\Edge[label=](131)(132)
	\Edge[label=](131)(133)
	\Edge[label=](131)(134)
	\Edge[label=](131)(135)
	\Edge[label=](131)(136)
	\Edge[label=](132)(133)
	\Edge[label=](132)(134)
	\Edge[label=](132)(135)
	\Edge[label=](132)(136)
	\Edge[label=](132)(137)
	\Edge[label=](133)(134)
	\Edge[label=](133)(135)
	\Edge[label=](133)(136)
	\Edge[label=](133)(137)
	\Edge[label=](133)(138)
	\Edge[label=](134)(135)
	\Edge[label=](134)(136)
	\Edge[label=](134)(137)
	\Edge[label=](134)(138)
	\Edge[label=](134)(139)
	\Edge[label=](135)(136)
	\Edge[label=](135)(137)
	\Edge[label=](135)(138)
	\Edge[label=](135)(139)
	\Edge[label=](135)(140)
	\Edge[label=](136)(137)
	\Edge[label=](136)(138)
	\Edge[label=](136)(139)
	\Edge[label=](136)(140)
	\Edge[label=](136)(141)
	\Edge[label=](137)(138)
	\Edge[label=](137)(139)
	\Edge[label=](137)(140)
	\Edge[label=](137)(141)
	\Edge[label=](137)(142)
	\Edge[label=](138)(139)
	\Edge[label=](138)(140)
	\Edge[label=](138)(141)
	\Edge[label=](138)(142)
	\Edge[label=](138)(143)
	\Edge[label=](139)(140)
	\Edge[label=](139)(141)
	\Edge[label=](139)(142)
	\Edge[label=](139)(143)
	\Edge[label=](139)(144)
	\Edge[label=](140)(141)
	\Edge[label=](140)(142)
	\Edge[label=](140)(143)
	\Edge[label=](140)(144)
	\Edge[label=](140)(145)
	\Edge[label=](141)(142)
	\Edge[label=](141)(143)
	\Edge[label=](141)(144)
	\Edge[label=](141)(145)
	\Edge[label=](141)(146)
	\Edge[label=](142)(143)
	\Edge[label=](142)(144)
	\Edge[label=](142)(145)
	\Edge[label=](142)(146)
	\Edge[label=](142)(147)
	\Edge[label=](143)(144)
	\Edge[label=](143)(145)
	\Edge[label=](143)(147)
	\Edge[label=](143)(148)
	\Edge[label=](144)(145)
	\Edge[label=](144)(146)
	\Edge[label=](144)(147)
	\Edge[label=](144)(148)
	\Edge[label=](144)(149)
	\Edge[label=](145)(146)
	\Edge[label=](145)(147)
	\Edge[label=](145)(149)
	\Edge[label=](146)(147)
	\Edge[label=](146)(148)
	\Edge[label=](146)(149)
	\Edge[label=](147)(148)
	\Edge[label=](147)(149)
	\Edge[label=](148)(149)
	\end{tikzpicture}}
\captionsetup{width=1\textwidth}
\caption{\textbf{(a)} Cumulative distribution of the shortest path distance approximation error on the PPI network~\cite{chatr2014ppi} ($\left | V \right |$=3,890) for embedding dimension $d$. The distortion for our method (D2V) is much smaller than state of the art (N2V). The distortion error for nodes $u$ and $v$ is defined as  $e_{u,v} =  \left|d(u,v) - \gamma \cdot \left\|\mathbf{X}_v-\mathbf{X}_u\right\|\right|/d(u,v)$.  Embeddings were created with parameter settings such that $n=10$,\, $\widehat{w}=10$,\, $\alpha = 0.025$,\, $k=1$, $l=40$ (D2V) and $l=80$ (N2V). The best inout and return parameters of N2V were chosen with grid search over $\{0.25,0.5,1,2,4\}$. \textbf{(b)} Visualization of a Watts-Strogatz graph with our embedding procedure.}

\end{figure}
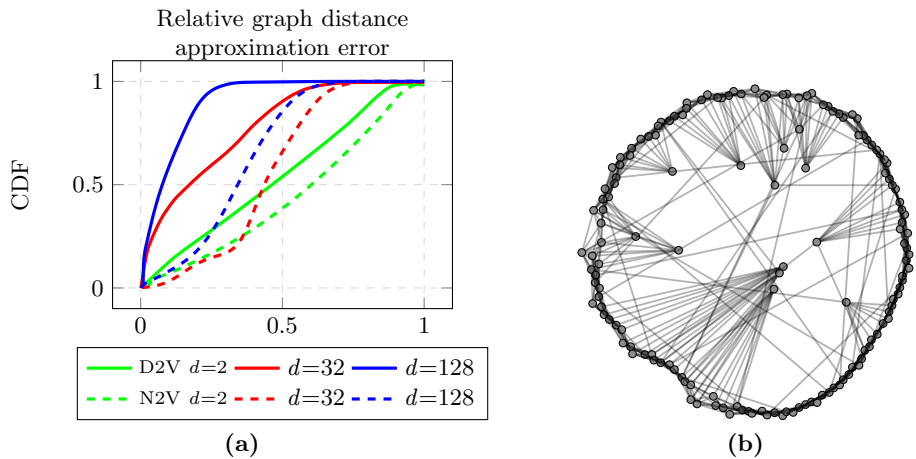
\vspace{-2em}

%% file: sections/related.tex
%!TEX root = sequence.tex

\section{Related Works}\label{sec:related}
Well known embedding techniques use a matrix that describes the graph and factorize it in order to create the embedding of the network. One can factorize the adjacency, neighbourhood overlap or Laplacian matrices. Based on the properties of the matrix either eigenvalue decomposition or some variant of stochastic gradient descent is used to obtain the graph embedding. These embedding methods all have a weakness, namely that they are computationally expensive. We refer the reader to the recent survey in~\cite{goyal2017graph} for a broader overview of graph embedding, and focus here on relevant neural network based embeddings. 

\myparagraph{Sequence based embedding.} The generation of node sequence based graph embeddings consists of three phases. First, the algorithm creates vertex sequences - usually by a random process. Second, features that are extracted from the synthetic sequences describe the approximated proximities of nodes. Finally, the embedding itself is learned using the extracted node specific features with a neural network which has a single hidden layer. Sequence based embedding originates from the \textit{DeepWalk} model \cite{perozzi2014deepwalk} which uses random walks to generate node sequences. This approach was improved upon by \textit{Node2Vec} (henceforth N2V) \cite{grover2016node2vec} which uses second-order random walks to generate the vertex sequences. Second-order random walks alternate between depth-first and breadth-first search on the graph in a random, but somewhat controlled way. In this attempt to have greater control on random walks, N2V introduces parameters that affect the embedding quality and are hard to optimize.

%% file: sections/theory.tex
%!TEX root = sequence.tex

\section{Feature extraction and neural network embedding}\label{sec:theory}

\myparagraph{Feature extraction.} We start with extracting features called hitting frequency vectors -- denoting frequencies with which vertices occur near each other.  The graph is denoted by $\mathcal{G}(V,E)$. The set of vertices is $V$ and the edge set is $E$. We assume that the graph is undirected and unweighted. Let us consider an example to see how an embedding is generated. 

Consider the example in Figure~\ref{fig:graph_example}(a). The vertex set contains nodes $a,b,c,d,e$ and nodes are indexed respectively from 1 to 5, and suppose we are given the 3 node sequences in the figure. To generate features from the sequences we choose a sliding window size denoted by $\widehat{w}$ which limits the maximal graph proximity among nodes that we are going to approximate.  In this case we choose $\widehat{w}=1$. We calculate the co-occurrence frequencies for node $c$ as follows -- we count how many times other nodes appeared at given positions before and after $c$ limited by the window's size. In this toy example it means positions at maximal 1 step before or after $c$ in the sequence. Counts at different positions are stored in separate vectors for each node. The resulting frequency vectors are as follows:
$\textbf{y}_{c,-1}=\begin{bmatrix}
1& 
1& 
0& 
2&
0 
\end{bmatrix}$ and $\textbf{y}_{c,+1}=\begin{bmatrix}
0& 
0&
0& 
4&
0
\end{bmatrix}$. Components of the vectors can be interpreted as noisy proximity statistics in the graph. The idea is that nearby nodes will have higher values in each-other's vectors. We concatenate these vectors to form a vector of $2\cdot \widehat{w} \cdot \left | V \right |$ components and call it the hitting frequency vector $\textbf{y}_v$ of a node $v$. We construct such a hitting frequency vector for each node from the given sequences. 
\input{./figures/neural_network.tex}
\myparagraph{Learning an embedding from the features.} For each vertex $v\in V$, we wish to compute a coordinate in $\reals^{d}$. The set of hitting frequency vectors is a representation of the graph in $\mathbb{R}^{|V|\times 2\cdot \widehat{w}\cdot |V|}$, which we have to reduce to a $\mathbb{R}^{|V|\times d}$ space. We write as $\textbf{x}_v$ the indicator (sometimes called hot-one) vector for $v$, which has $\abs{V}$ elements, all of which are zero, except the element at index of $v$, which is set to $1$. A schematic of the neural network architecture is in Figure~\ref{fig:graph_example}(b). The neural network has $d$ hidden neurons, each with $\abs{V}$ inputs and $2\cdot\widehat{w}\cdot\abs{V}$ outputs. The incoming and outgoing weight matrices of the hidden neurons are written as $\textbf{W}_{in}$ and $\textbf{W}_{out}$. To train the neural network, the training algorithm uses input output pairs of the form $(\textbf{x}_v,\textbf{y}_v)$ corresponding to each vertex $v$. Thus, the neural network learns to associate with each vertex, an output that is its hitting frequency vector. After the training, the incoming weight matrix $\textbf{W}_{in}$ (of dimension $d\times \abs{V}$) gives the $d$ dimensional embedding of the vertices.

 The weight matrix is used to approximately reconstruct the hitting frequencies of a node. If two nodes have similar hitting frequency vectors, meaning that their proximity is high, they will also have a similar latent space representation. Our goal is the efficient and scalable learning of the embedding so we use asynchronous gradient descent (ASGD). Analogous to previous works~
  \cite{perozzi2014deepwalk,grover2016node2vec}, we used hierarchical softmax activation with multinomial logloss, with which the computational complexity of a training epoch (while we decrease the learning rate from starting value to zero) is $\mathcal{O}(|V|\log (|V|))$. We refer to the embedding as $\textbf{X}$,  and the embedding of node $v$ is noted by $\textbf{X}_v$.

\vspace{-1.5em}

%% file: figures/neural_network.tex
\vspace{-3em}
\begin{figure}[H]
\centering
\captionsetup[subfloat]{width=140pt}

	\subfloat[]{
		\scalebox{1}{
			
			\begin{tikzpicture}[transform shape]
			\SetVertexNormal[Shape      = circle,
			MinSize    =10pt,
			]
			\Vertex[L=$a$,x=0,y=2]{1}
			\Vertex[L=$b$,x=0,y=0]{2}
			\Vertex[L=$c$,x=2,y=0]{3}
			\Vertex[L=$d$,x=2,y=2]{4}
			\Vertex[L=$e$,x=4,y=2]{5}
					\tikzstyle{VertexStyle} = [shape = rectangle]
			\Vertex[L=$a-b-c-d-c-d-e$,x=2,y=-0.75]{11}
			\Vertex[L=$e-d-e-d-c-d-e$,x=2,y=-1.5]{12}
			\Vertex[L=$b-a-c-d-a-b-a$,x=2,y=-2.25]{13}
			
			\tikzstyle{VertexStyle} = [shape = rectangle]
			
			\tikzstyle{EdgeStyle}=[]
			\Edge[label=](1)(2)
			\Edge[label=](1)(4)
			\Edge[label=](2)(3)
			\Edge[label=](3)(1)
			\Edge[label=](3)(4)	
			\Edge[label=](4)(5)	
			\Edge[label=](3)(5)
			\end{tikzpicture}}}
	\quad\quad
	\subfloat[]{\scalebox{0.8}{
	\begin{tikzpicture}[transform shape]
	\SetVertexNormal[Shape      = circle,
	MinSize    =3.5pt,
	]
	\Vertex[L=$$,x=-2,y=2]{1}
	\Vertex[L=$$,x=-1,y=2]{2}
	\Vertex[L=$$,x=0,y=2]{3}
	\Vertex[L=$$,x=1,y=2]{4}
	\Vertex[L=$$,x=2,y=2]{5}
	
	\Vertex[L=$$,x=-0.75,y=4]{6}
	\Vertex[L=$$,x=0.75,y=4]{7}

	\Vertex[L=$$,x=-2.5,y=6]{10}
	\Vertex[L=$$,x=-2,y=6]{11}	
	\Vertex[L=$$,x=-1.5,y=6]{12}
	\Vertex[L=$$,x=-1,y=6]{13}		
    \Vertex[L=$$,x=-0.5,y=6]{14}

    \Vertex[L=$$,x=2.5,y=6]{30}
    \Vertex[L=$$,x=2,y=6]{31}	
    \Vertex[L=$$,x=1.5,y=6]{32}
    \Vertex[L=$$,x=1,y=6]{33}		
    \Vertex[L=$$,x=0.5,y=6]{34}

\tikzstyle{VertexStyle} = [shape = rectangle]

\Vertex[L={$\textbf{h}_v=\sigma(\textbf{W}_{in}\cdot\textbf{x}_v+\textbf{b}_{in})$},x=3,y=4]{900}

\Vertex[L={$\widehat{\textbf{y}}_{v-1}$},x=-1.5,y=6.75]{1000} 
\Vertex[L={$\widehat{\textbf{y}}_{v+1}$},x=1.5,y=6.75]{1100} 
\Vertex[L={$\widehat{\textbf{y}}_v=\Phi(\textbf{W}_{out}\cdot\textbf{h}_v+\textbf{b}_{out})$},x=0,y=7.5]{1200}
\Vertex[L={$\textbf{x}_v$},x=0,y=1]{1300}

	\tikzstyle{EdgeStyle}=[bend left]
	\tikzstyle{LabelStyle}=[fill=white]

	\tikzstyle{EdgeStyle}=[post,opacity = 0.25]
	\Edge[label=](1)(6)
	\Edge[label=](2)(6)
	\Edge[label=](3)(6)
	\Edge[label=](4)(6)
	\Edge[label=](5)(6)
	\Edge[label=](1)(7)
	\Edge[label=](2)(7)
	\Edge[label=](3)(7)
	\Edge[label=](4)(7)
	\Edge[label=](5)(7)
	\tikzstyle{EdgeStyle}=[post,opacity = 0.25]
	\Edge[label=](6)(10)
	\Edge[label=](6)(11)
	\Edge[label=](6)(12)
	\Edge[label=](6)(13)
	\Edge[label=](6)(14)	

	\Edge[label=](7)(10)
	\Edge[label=](7)(11)
	\Edge[label=](7)(12)
	\Edge[label=](7)(13)
	\Edge[label=](7)(14)

	\Edge[label=](6)(30)
	\Edge[label=](6)(31)
	\Edge[label=](6)(32)
	\Edge[label=](6)(33)
	\Edge[label=](6)(34)	
	
	\Edge[label=](7)(30)
	\Edge[label=](7)(31)
	\Edge[label=](7)(32)
	\Edge[label=](7)(33)
	\Edge[label=](7)(34)

	\draw [thick,decoration={brace,mirror},
	yshift=-10pt,
	decorate
	](-2.3,1.95) -- (2.3,1.95) node [midway,yshift=-0.55cm] {}; 
	
	\draw [thick,decoration={brace},yshift=-10pt,decorate](-2.7,6.7) -- (-0.3,6.7) node [midway,yshift=0.35cm] {}; 
	\draw [thick,decoration={brace},yshift=-10pt,decorate](0.3,6.7) -- (2.7,6.7) node [midway,yshift=0.35cm] {}; 

	\draw [thick,decoration={brace},yshift=-10pt,decorate](-2.7,7.45) -- (2.7,7.45) node [midway,yshift=0.35cm] {}; 
\end{tikzpicture}}}
\captionsetup{width=1\textwidth}
\caption{\textbf{(a)} Graph with linear vertex sequences. The three vertex sequences listed are used for feature extraction in our example. \textbf{(b)} Architecture of the example neural network.}\label{fig:graph_example}
\end{figure}
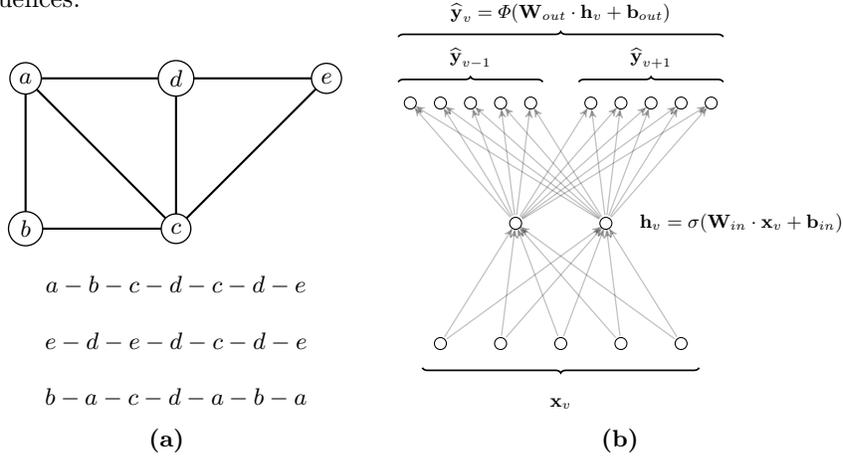
\vspace{-2em}

%% file: sections/algorithm.tex
\section{Sequence generation algorithm and design}\label{sec:algorithm}
\vspace{-0.5em}
\begin{minipage}{\linewidth}
	\begin{minipage}[t]{6.5cm}
		\vspace{0pt}  
		\input{./algorithms/euler_sampler.tex}
		$$$$
		\input{./algorithms/sequence_generation.tex}		
	\end{minipage}%
	\hspace{-30pt}
	\begin{minipage}[t]{6.5cm}
		\vspace{0pt} 
To generate sequences in the neighborhood of a node, we first compute a \textit{diffusion graph}, and then use it to compute vertex sequences. 
		
\myparagraph{Diffusion graph generation.} We emulate a simple diffusion-like random process starting from a vertex $v$ to sample a subgraph of $l$ vertices near $v$. The diffusion graph $\widetilde{\mathcal{G}}$ is initialized with $\{v\}$. Next, at each step, we sample a random node $u$ from $\widetilde{\mathcal{G}}$ and from the neighbors of $u$ in the original graph $\mathcal{G}$, we select $w$. We add $w$ to the set of vertices in $\widetilde{\mathcal{G}}$, and add the edge $(u,w)$ to $\widetilde{\mathcal{G}}$. This process is repeated until $\widetilde{\mathcal{G}}$ has $l$ nodes.

\myparagraph{Node sequence sampling.} To generate sequences from the subgraph $\widetilde{\mathcal{G}}$, we take the following approach. We convert $\widetilde{\mathcal{G}}$ into a multigraph by doubling each edge into two edges. A connected graph where every node has an even degree is Eulerian, and the Euler walk is easy to find~\cite{west2001introduction}. We use this method to find the Euler walk and use that as a vertex sequence. Observe that this diffusion graph sampling and sequence generation can be performed in parallel across many machines, since each diffusion graph can be generated independent of others. The generated sequences are then used to produce graph embedding using neural networks as seen in the previous section. Note that an Euler walk has the nice property that it captures every adjacency relation in the subgraph into a linear sequence using asymptotically optimal space. This property then helps our method perform better both in the sense of efficiency and quality of results. 
	\end{minipage}%
	
\end{minipage}

%% file: algorithms/euler_sampler.tex
\scalebox{0.75}{{\small\begin{algorithm}[H]
	\vspace{2mm}
\DontPrintSemicolon
\SetAlgoLined
\footnotesize
\KwData{$\mathcal{G}$ -- Graph object.\\
\quad \quad\,\,\,\,\,\,$l$ -- Number of nodes sampled.\\
\quad \quad\,\,\,\,\,\,$v$ -- Starting node .}
\KwResult{$P$ -- Eulerian sequence from $v$.}
\vspace{2mm}
%$\widetilde{\mathcal{G}}\leftarrow K_0$\; 
$V_{\widetilde{\mathcal{G}}} \leftarrow \left\{ v  \right \}$\;
\While{$ |V_{\widetilde{\mathcal{G}}}| < l$}{
	\vspace{2mm}
 	$w\leftarrow \text{Random Sample}(V_{\widetilde{\mathcal{G}}})$\;
 	$u\leftarrow \text{Random Sample}(N_{\mathcal{G}}\left(w\right))$\;

		\If{$ u \notin  V_{\widetilde{\mathcal{G}}}$}{	
			$V_{\widetilde{\mathcal{G}}} \leftarrow V_{\widetilde{\mathcal{G}}}\cup \left \{ u  \right \}$\;
			$E_{\widetilde{\mathcal{G}}} \leftarrow E_{\widetilde{\mathcal{G}}}\cup \left \{( u,w) \right \}$\;
		}

}
$\widetilde{\mathcal{G}} \leftarrow \text{Duplicate Edges}(\widetilde{\mathcal{G}})$\;
$P \leftarrow \text{Random Eulerian Circuit}(\widetilde{\mathcal{G}},v)$\;
\vspace{2mm}
\caption{Graph sampling}\label{eulerian_diffusion}
\end{algorithm}

}}

%% file: algorithms/sequence_generation.tex
\scalebox{0.75}{{\small\begin{algorithm}[H]
\vspace{2mm}
\DontPrintSemicolon
\SetAlgoLined
\footnotesize
\KwData{$\mathcal{G}$ -- Graph embedded.\\
\quad \quad\,\,\,\,\,\,$p$ -- Sequence samples per node.\\
\quad \quad\,\,\,\,\,\,$l$ -- Number of nodes per sample.\\
\quad \quad\,\,\,\,\,\,$d$ -- Dimension of embedding.\\
\quad \quad\,\,\,\,\,\,$k$ -- Number of epochs.\\
\quad \quad\,\,\,\,\,\,$\widehat{w}$ -- Size of sliding window.\\
\quad \quad\,\,\,\,\,\,$\alpha$ -- Learning rate.}
\KwResult{$\textbf{X}$ -- Embedding of graph $\mathcal{G}$.}
\vspace{2mm}
$ \mathcal{G}_{1},\dots \mathcal{G}_{S}\leftarrow\text{Component Extraction}(\mathcal{G})$\; 
$\text{Samples}\leftarrow []$\;
\For{$i \,\, \text{\upshape in} \,\, 1:p$}{
    \vspace{2mm}
	$\text{Walks}\leftarrow \{ \}$\;
	$l' \leftarrow l$\;
    \For{j\,\,\text{\upshape in}\,\,1:$\left| \left\{ \mathcal{G}_{1},\mathcal{G}_{2},\dots \mathcal{G}_{S}\right\} \right |$}{
	    \vspace{2mm}
		\If{$\left | V_{\mathcal{G}_j} \right |<l'$}{	
			$ l' \leftarrow \left | V_{\mathcal{G}_j} \right |$\;		
		}
		\For{$v\,\, \text{\upshape in}\,\, V$}{	
			$\text{Walks}(v) \leftarrow\text{Traceback}(\mathcal{G}_j,v,l') $\;
		}
     }
$\text{Samples}(i)\leftarrow \text{Walks}$\;
}
    $\textbf{X}\leftarrow \text{Learn Emb.}(\text{Samples},d,\widehat{w},\alpha,k)$\;
    \vspace{2mm}
    \caption{Learning from sequences}\label{alg:sequence_generation}
\end{algorithm}
}}

%% file: sections/experiment.tex
%!TEX root = sequence.tex

\section{Experiments}\label{sec:experiments}
In our experiments we compare our method D2V with the state of the art N2V~\cite{grover2016node2vec} method. We look at quality of embeddings and the computational performance. The main observations from the experiments are:
\begin{itemize}
	\item With increasing size of graphs, efficiency of D2V scales better than that of N2V. 
	\item The D2V embedding preserves distances well between most pairs of nodes: in 128 dimensional embedding, over 90\% pairs suffer a distortion smaller than 20\%. In any dimensions, it performs bettern than N2V. 
	\item Clustering of the D2V embedding works well for community detection, and performs better than N2V measured by the modularity of clusters.
\end{itemize}
\vspace{-2em}
\input{./tables/timing.tex}

\myparagraph{Computational efficiency.} In the first series of experiments we measured the average graph pre-processing and sequence generation times on a number of real world networks. Pre-processing in this case involves reading the graph and creating suitable data structures.  N2V in particular requires data structures to regularly update the random walk probabilities. Note that it is the preprocessing and sequence generation where these two methods differ, as they use similar methods for training neural networks. Our results in Table~\ref{tab:timing} show that on larger networks D2V has a consistent advantage performance wise.

\myparagraph{Node distance approximation.} Using the PPI network we measure how well the shortest path distance of nodes $d(u,v)$ can be approximated by the Euclidean distance of nodes in the embedding space. The relative approximation error $e_{u,v}$ for a given pair of nodes $u,v$ is defined by $ e_{u,v} =  \left|d(u,v) - \gamma \cdot \left\|\mathbf{X}_v-\mathbf{X}_u\right\|\right|/d(u,v)$. Essentially, the absolute difference between $d(u,v)$ and the scaled Euclidean distance to $d(u,v)$. The factor $\gamma$ adjusts for the uniform scaling over the graph. We take the $\gamma$ that minimises the sum of errors. 

We plotted cumulative distribution of the relative approximation error for different embedding dimensions on Figure \ref{fig:approxcdfs}. With a 32 dimensional D2V embedding one can approximate half of the shortest path distances with a relative error below 20\%. Increasing the embedding dimension to 128 allows to approximate 90\% of shortest paths with an approximation error below 20\%. Finally, we also plotted the approximation error obtained with N2V embeddings. A 32 dimensional N2V embedding can only approximate roughly 10\% of the shortest path distances with a relative error below 20\%. Moreover, increasing the N2V embedding dimension does not decrease the distortion considerably.  We conclude that on this graph D2V approximates graph distances better than N2V.
 \myparagraph{Community detection.} We evaluated the utility of the embedding in community detection. We clustered the embedded nodes in the embedding space using k-means clustering, and then computed modularity~\cite{newman2006modularity} as a quality measure.  The experiments involved six different datasets with number of vertices ranging from few thousands to millions and we compared our results to clusterings obtained with standard community detection methods. Results are seen in Table~\ref{tab:cluster_results}. Our results show that k-means clustering of the embeddings outperforms all other methods on most of the datasets. Moreover, D2V (our method)  results in clusterings that are higher quality than clusters created with N2V.
\vspace{-1.5em}
 \input{./tables/clustering.tex} 
\vspace{-0.1em}

%% file: tables/timing.tex
\begin{table}[h!]
	\centering
\begin{tabular}{l cccccc}
\toprule
&\multicolumn{2}{c}{\textbf{\textsc{Blogcatalog}}}&\multicolumn{2}{c}{\textbf{\textsc{\quad \quad PPI \quad  \quad}}}&\multicolumn{2}{c}{\textbf{\textsc{\quad Wikipedia \quad}}}\\
&\multicolumn{2}{c}{\scriptsize{$\left | V \right |$=10,312}}&\multicolumn{2}{c}{\scriptsize{$\left | V \right |$=3,890}}&\multicolumn{2}{c}{\scriptsize{$\left | V \right |$=4,777}}\\
&\multicolumn{2}{c}{\scriptsize{$\left | E \right |$=333,982}}&\multicolumn{2}{c}{\scriptsize{$\left | E \right |$=38,739}}&\multicolumn{2}{c}{\scriptsize{$\left | E \right |$=92,517}}\\
\cmidrule(l){2-3}\cmidrule(l){4-5}\cmidrule(l){6-7}
&N2V&D2V&N2V&D2V&N2V&D2V\\

\hline

\textbf{Sequence generation}& 59.089  &\textbf{19.983} &  \textbf{4.253}  & 4.684 & 12.135  & \textbf{6.879}  \\[0.6em]
\textbf{Pre-processing}
  & 784.899 & \textbf{3.231} 
 & 12.797 & \textbf{0.362} 
& 185.287 & \textbf{0.667}  \\[0.6em]

\hline
\end{tabular}

\caption{Computation time on real life graphs. \textbf{BlogCatalog:} Is a social network of bloggers, nodes are bloggers and links are social relationships~\cite{agarwal2009blogcatalog}. \textbf{PPI:} is a protein-protein interaction network of humans~\cite{chatr2014ppi}. \textbf{Wikipedia:} Is a word co-occurrence network based on a chunk of the Wikipedia corpus \cite{mahoney2011wikipedia}. Columns report running time in seconds extracted from 100 experiments on the datasets. Bold numbers mark the fastest mean pre-processing -- sequence generation times on a given dataset. \label{tab:timing}}
\end{table}
\vspace{-2em}

%% file: tables/clustering.tex
\begin{table}[htbp]
	\centering
	
	\begin{tabular}{lcccccc}\toprule
\multicolumn{1}{c}{\textsc{\textbf{Algorithm}}}&\multicolumn{1}{c}{\textbf{\textsc{Blogcatalog}}}&\multicolumn{1}{c}{\textbf{\textsc{PPI}}} &\multicolumn{1}{c}{\textbf{\textsc{Wikipedia}}} &\multicolumn{1}{c}{\textbf{\textsc{Flickr}}}&\multicolumn{1}{c}{\textbf{\textsc{Youtube}}}&\multicolumn{1}{c}{\textbf{\textsc{Markercafe}}}  \\\toprule
		\textbf{Fast Greedy}         & 0.2069      & 0.3029 & \textbf{0.1456} &0.4517& --&0.2597  \\[0.2em]
		\textbf{Walktrap}             & 0.1766      & 0.2571 & 0.0553 &0.4873& --& 0.2026 \\[0.2em]
		\textbf{Eigenvector} & 0.2035      & 0.2262 & 0.0915 &0.4810& --& 0.2455 \\[0.2em]
		\specialrule{.1em}{.05em}{.05em} \\[-1ex]
		\textbf{K-means D2V}        & \textbf{0.2225}      & \textbf{0.3365} & 0.1420   &\textbf{0.5078}&\textbf{0.6265}&\textbf{0.2818}  \\[0.2em]
		\textbf{K-means N2V}      & 0.2184      & 0.3270  & 0.1376   &0.3647&0.4862&0.2630 \\[0.2em]
		\specialrule{.1em}{.05em}{.05em} \\[-1ex]
	%	\textbf{Hierarchical D2V}           & 0.1610       & 0.2618 & 0.0696 &--&--& --  \\[0.3em]
	%	\textbf{Hierarchical N2V}          & 0.1149      & 0.2774 & 0.0709 &--&-- &--\\[0.3em]
%		\specialrule{.1em}{.05em}{.05em} \\[-1ex]
	\end{tabular}

\caption{Clustering quality measured by modularity. The baseline community detection algorithms can be found in \cite{clauset2004finding,pons2006computing,newman2006modularity}.
	Bold numbers note the highest modularity value obtained on the dataset. Dashes denote missing modularity values when obtaining a clustering was not feasible due to computational complexity of the algorithm. Embeddings were created with baseline parameter settings such that $d=128$,\, $n=10$,\, $\widehat{w}=10$,\, $\alpha = 0.025$,\, $k=1$, $l=40$ (D2V) and $l=80$ (N2V). The best input and return parameters of N2V were chosen with grid search over $\{0.25,0.5,1,2,4\}$ while the cluster number varied between 2 and 50. The distance measure was the Euclidean distance in the latent space. Besides the earlier used datasets we chose 3 additional social networks to asses the representation quality. \textbf{Flickr:} A network of Flickr users \cite{mcauely2012flickr}.  \textbf{Youtube:} Is a friendship network of Youtube users \cite{leskovec2015youtube}. \textbf{Markercafe:} Is data from an Israeli social network \cite{fire2011markercafe}.}\label{tab:cluster_results}
\end{table}

%% file: sections/conclusions.tex
\vspace{-3.5em}
\section{Conclusions}\label{sec:conclusions}
In this work we proposed \textit{Diff2Vec} a node sequence based graph embedding model that uses diffusion processes on graphs to create vertex sequences. We demonstrated that the design of the algorithm results in fast sequence creation in realistic settings. It also allows parallel vertex sequence generation which leads to additional speed up. We confirmed that node features created with \textit{Diff2Vec} are useful features for downstream machine learning tasks. We gave a detailed evaluation of the representation quality of embeddings on shortest path distance approximation and the machine learning task of community detection. Our findings show that besides the favourable computational performance the representation quality itself is competitive with other methods.\\

\noindent \textbf{Acknowledgements:} Benedek Rozemberczki was supported by the Centre for Doctoral Training in Data Science, funded by EPSRC (grant EP/L016427/1).
\bibliographystyle{plain}
\bibliography{main}